\documentclass[conference, anonymous]{IEEEtran}
\usepackage{cite}
\usepackage{amsmath,amssymb,amsfonts}
\usepackage{algorithmic}
\usepackage{graphicx}
\usepackage{textcomp}
\usepackage{xcolor}
\usepackage{cite}
\usepackage{amsmath,amssymb,amsfonts}
\usepackage{algorithmic}
\usepackage[linesnumbered,ruled,vlined]{algorithm2e}
\usepackage{comment}
\usepackage{blindtext}
\usepackage{siunitx}
\usepackage{graphicx}
\usepackage{subcaption}
\usepackage{balance}
\usepackage{bm}
\usepackage{multirow}
\usepackage{booktabs}
\usepackage{hyperref}
\usepackage{url}
\usepackage{threeparttable}

\newcommand{\solution}{\textit{RAG-HAR{}}}
\makeatletter
\newcommand{\linebreakand}{
  \end{@IEEEauthorhalign}
  \hfill\hbox{}\par
  \mbox{}\hfill\begin{@IEEEauthorhalign}
}
\makeatother

\def\BibTeX{{\rm B\kern-.05em{\sc i\kern-.025em b}\kern-.08em
    T\kern-.1667em\lower.7ex\hbox{E}\kern-.125emX}}
\begin{document}

\title{RAG-HAR: Retrieval Augmented Generation-based Human Activity Recognition
}


\author{
\IEEEauthorblockN{Nirhoshan Sivaroopan\IEEEauthorrefmark{1}}
\IEEEauthorblockA{\textit{University of Sydney} \\
Australia }
\and
\IEEEauthorblockN{Hansi Karunarathna\IEEEauthorrefmark{1}}
\IEEEauthorblockA{\textit{University of Sri Jayewardenepura} \\
Sri Lanka}
\and
\IEEEauthorblockN{Chamara Madarasingha}
\IEEEauthorblockA{\textit{Curtin University} \\
Australia }
\linebreakand 
\IEEEauthorblockN{Anura Jayasumana}
\IEEEauthorblockA{\textit{Colorado State University} \\
USA }
\and
\IEEEauthorblockN{Kanchana Thilakarathna}
\IEEEauthorblockA{\textit{University of Sydney} \\
Australia }
}

\maketitle
\begin{abstract}
  Human Activity Recognition (HAR) underpins applications in healthcare, rehabilitation, fitness tracking, and smart environments, yet existing deep learning approaches demand dataset-specific training, large labeled corpora, and significant computational resources. We introduce \solution{}, a training-free retrieval-augmented framework that leverages large language models (LLMs) for HAR. \solution{} computes lightweight statistical descriptors, retrieves semantically similar samples from a vector database, and uses this contextual evidence to make LLM based activity identification. We further enhance \solution{} by first applying prompt optimization and introducing an LLM-based activity descriptor that generates context-enriched vector databases for delivering accurate and highly relevant contextual information. Along with these mechanisms, \solution{} achieves state-of-the-art performance across six diverse HAR benchmarks. Most importantly, \solution{} attains these improvements without requiring model training or fine-tuning, emphasizing its robustness and practical applicability. \solution{} moves beyond known behaviors, enabling the recognition and meaningful labelling of multiple unseen human activities. 
\end{abstract}


\section{Introduction}

Human Activity Recognition (HAR) from wearable sensor data enables continuous monitoring, anomaly detection, and personalized interventions across healthcare~\cite{bachlin2009wearable}, rehabilitation~\cite{panwar2019rehab}, fitness~\cite{liaqat2019wearbreathing}, and smart environments~\cite{grzeszick2017deep}. Despite wide-ranging applications, HAR remains challenging due to inter-subject variability, differences in sensor placement, device heterogeneity, and subtle distinctions between activities that exhibit similar motion patterns~\cite{suh2022adversarial}. Those challenges create a strong need for accurate, generalizable, and cost-efficient solutions.

Deep learning (DL) has become the dominant paradigm for HAR, with convolutional neural networks (CNNs)~\cite{zeng2014convolutional, bhattacharya2016sparsification}, recurrent architectures~\cite{guan2017ensembles, hammerla2016deep}, and attention-based models~\cite{ahmad2023alae} achieving state-of-the-art (SOTA) performance on benchmark datasets. However, DL-based HAR faces three critical limitations: (i) costly and time-consuming training procedures tailored to each dataset; (ii) performance degradation under domain shift across subjects, sensor placements, or devices; and (iii) heavy dependence on large labeled datasets~\cite{chen2020metier, sheng2020weakly}. Transfer learning~\cite{wang2018stratified, haresamudram2023investigating} and data augmentation~\cite{ahmad2023alae} partially mitigate these issues but continue to rely on resource-intensive training pipelines. Despite advances in DL, these limitations leave HAR without a practical solution that is simultaneously training-free, generalizable, and scalable.

\begin{figure}[t]
    \centering
    \includegraphics[width=\columnwidth]{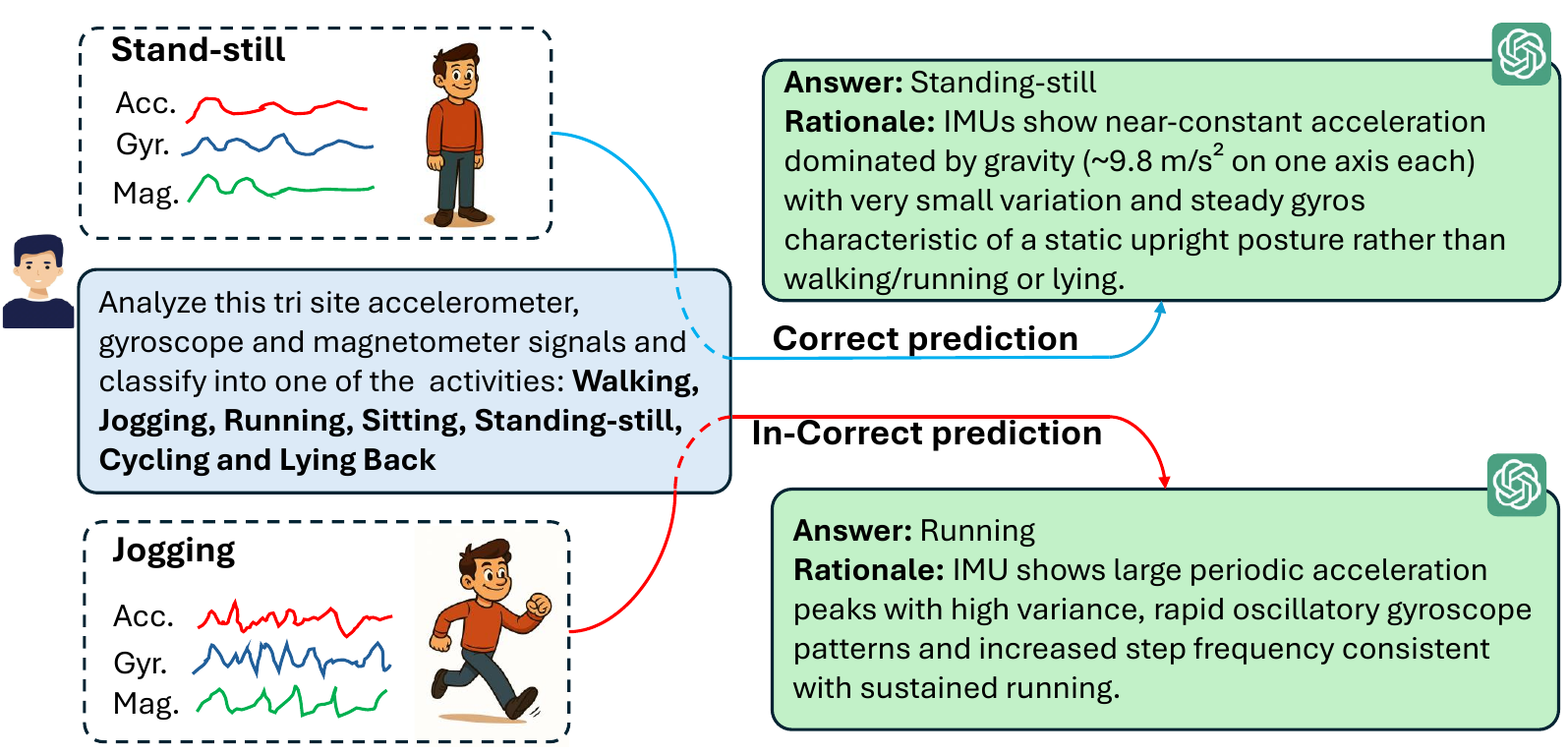} 
    \caption{Success and failure cases of LLMs in HAR.}
    \label{fig:llm-failures}
\end{figure}

To address this gap, this paper explores a fundamentally different paradigm: leveraging Large Language Models (LLMs) as reasoning engines for HAR. LLMs pretrained on massive and diverse corpora offer three properties that support this direction: (i) broad world knowledge that includes textual descriptions of human activities; (ii) strong pattern-matching abilities that support cross-domain generalization; and (iii) reasoning capabilities suitable for complex, context-dependent tasks. 
These properties suggest that LLMs could be applied beyond language to structured sensor data reasoning tasks. However, when directly applied to wearable sensor signals, LLMs often misclassify activities for two main reasons. First, as shown in Fig.~\ref{fig:llm-failures}, they perform well on activities with distinct motion signatures but struggle with fine-grained cases that rely on subtle discriminative cues. Second, these models lack grounding in activity-specific knowledge, which is essential for contextual awareness and accurate predictions\cite{li2024sensorllm}.


These limitations motivate our central idea: \emph{Augmenting LLMs with retrieval of semantically relevant activity examples while capturing subtle variations of activity patterns through calculated statistical features}. Thus, we introduce \textbf{\solution{}}, a retrieval-augmented framework that eliminates the need for task-specific classifier training. We establish a novel perspective for the HAR community: shifting from \emph{model-centric learning} to \emph{\textbf{retrieval-augmented reasoning}}.

\solution{} implements a two-stage framework. In the \textbf{Baseline stage}, sensor data streams are partitioned into sliding windows, from which statistical descriptors are extracted and embedded to form a vectorized knowledge base. During inference, the system retrieves the top-$k$ nearest neighbors from the vector database and provides them, together with the query descriptors, as context to the LLM for classification.
By grounding predictions in relevant prior examples, the baseline stage achieves SOTA performance across five benchmark HAR datasets—\emph{without any dataset-specific training}—while remaining simple, fast, and cost-effective.  
\textbf{Optimization stage} builds on the baseline with two enhancement strategies: (1) \emph{retriever enhancement}, where LLMs generate richer, semantically meaningful descriptors of time-series windows to improve semantic matching; and (2) \emph{prompt optimization}, which employs a search-based optimizer to systematically explore the prompt space and identify effective strategies and phrasings that maximize the LLM classification performance. These enhancements improve accuracy but require additional calibration and more LLM invocations, offering a controllable trade-off between efficiency and performance.

The two-stage framework offers multiple advantages. First, compared to SOTA DL approaches, the framework eliminates dataset-specific training and sharply reduces computational overhead and deployment latency. Second, maintenance of heavyweight DL models is unnecessary: inference relies on lightweight retrieval and LLM calls rather than local model hosting. Third, the database-centric design enables natural scalability, as new activity examples can be added to the database, whereas DL models typically require retraining. Fourth, adoption barriers decrease because deployment does not require extensive ML/DL expertise, specialized GPU infrastructure, or complex dataset augmentation pipelines. Finally, and \textbf{most importantly}, unlike any existing ML/DL framework, \solution{} can classify multiple previously unseen activity classes and even generate meaningful labels for these novel activities—capabilities that traditionally require extensive human expertise in activity annotation and time-series analysis. This unprecedented ability to handle unseen classes without retraining or manual labeling \textbf{fundamentally distinguishes \solution{} from all prior approaches. }

\noindent
The main contributions of this paper are as follows:
\begin{itemize}
    \item We introduce \solution{}, the first retrieval-augmented framework for HAR that attains SOTA performance without any model training. Across six benchmark datasets, the framework improves F1-Score over leading DL baselines by approximately 0.5\% - 6.2\% while maintaining a lightweight inference pipeline.
    \item We design a two-stage framework that introduces two novel enhancement strategies; enhanced retrieval descriptors for more precise semantic matching, and a prompt optimizer that systematically explores the prompt space to identify effective instructions tailored to each dataset. Together, these strategies explicitly optimize the retriever and the LLM while offering complementary accuracy gains. This design makes \solution{} not only a fast and cost-effective baseline for training-free HAR but also a flexible framework that adapts to different deployment scenarios and resource budgets by trading efficiency for accuracy.
    \item We provide an extensive empirical evaluation on six diverse datasets, covering sensor channels from 6 to 60 and activity types that span locomotion, daily living, exercise, assembly-line tasks, and fine-grained variations in direction, pace, or object use. Experiments include both seen-subject and unseen-subject splits, demonstrating consistent gains over prior work.
    \item We demonstrate, for the first time, that LLMs can move beyond fixed taxonomies in HAR by detecting unseen activities and generating semantically coherent labels. This supports open-world HAR, where novel behaviors are recognized and meaningfully categorized rather than collapsed into a single ``unknown'' class.
\end{itemize}

\section{Related Work}

\subsection{Background of HAR}
Sensor-based HAR is typically formulated as a time-series classification problem using data from IMUs and other devices. 
Early DL approaches replaced traditional ML pipelines with CNNs to extract local temporal features, but these struggled with long-range dependencies~\cite{zeng2014convolutional, bhattacharya2016sparsification}. RNNs, were then adopted to better capture short and long-term relationships ~\cite{guan2017ensembles, hammerla2016deep}. However, these models required large-scale labelled datasets. In parallel, deep metric learning was explored as an alternative, where activities are embedded in a manifold space using pairwise or triplet losses~\cite{khaertdinov2021deep, hermans2017defense}, but its effectiveness hinges on careful hard-sample selection and is degraded by outliers. Other research has focused on improving generalization across individuals, as activity execution varies widely. Transfer learning and domain adaptation approaches pre-train on source datasets and fine-tune on target ones~\cite{ zhao2020local, haresamudram2023investigating}, though their success depends on source–target similarity. Adversarial methods extract user-independent features~\cite{suh2022adversarial} but are unstable and resource-intensive. Another challenge is class imbalance, which skews model performance. To address this, augmentation techniques such as axis rotation and jittering have been applied ~\cite{faridee2019augtoact, um2017data}, along with mixup-based methods that interpolate between samples~\cite{abedin2021attend, ahmad2023alae}. Generative models such as GANs further synthesize realistic activity data~\cite{bai2020adversarial, soleimani2021cross}, though at the cost of increased training complexity. 

\subsection{LLMs with HAR}

LLMs have recently been adapted for time-series analysis~\cite{chen2025large, zhou2025enhancing, jang2024time}, and the HAR community has begun exploring their potential. Their ability to leverage world knowledge and reasoning makes them well-suited to bridge raw sensor signals with high-level semantic understanding of human activities. Recent frameworks highlight this promise: SensorLLM~\cite{li2024sensorllm} aligns sensor data with language through channel-specific tokens before task-aware tuning, while LLM4HAR~\cite{hong2025llm4har} integrates sensor adaptation, knowledge injection, and efficiency modules. PH-LLM~\cite{khasentino2025personal}, a LLM fine-tuned on expert case studies and wearable sensor data to provide personalized sleep and fitness coaching. SimCLR~\cite{chen2020simple}-finetuned encoders have also been combined with LLMs for HAR, where the LLM computes the Euclidean distance between encoded test samples and reference encodings from SimCLR and demonstrated for binary classification only \cite{hota2025evaluating}. LLMs have also shown value in related applications such as user recognition ~\cite{chen2024towards} and activity event descriptors ~\cite{post2025contextllm,civitarese2025large}.
However, existing methods often rely on fine-tuning LLMs for sensor sequences, which enhances sequence modelling but diminishes their broader real-world knowledge. However, HARGPT~\cite{ji2024hargpt} explored providing HAR signals directly to the LLMs and prompting to identify the activity, but  as shown in Fig.~\ref{fig:llm-failures}, the method only works for trivial activities and a small number of channels. When the number of sensor channels increases, HARGPT fails heavily, as LLMs have not been previously trained to work with a series of vectors.

\textit{In contrast to prior work, \solution{} avoids extensive model training such as in DL by supplying context directly to the LLM. Moreover, through semantic context retrieval, it overcomes the limitations such as in few-shot prompting (e.g., shorter context window and higher inference cost). To the best of our knowledge, this is the first training-free framework capable of handling multi-channel, multi-subject, and fine-grained HAR while preserving strong performance.}


\begin{figure*}[ht]
    \centering
    \begin{subfigure}{.40\textwidth}
        \centering
        \includegraphics[width=\linewidth]{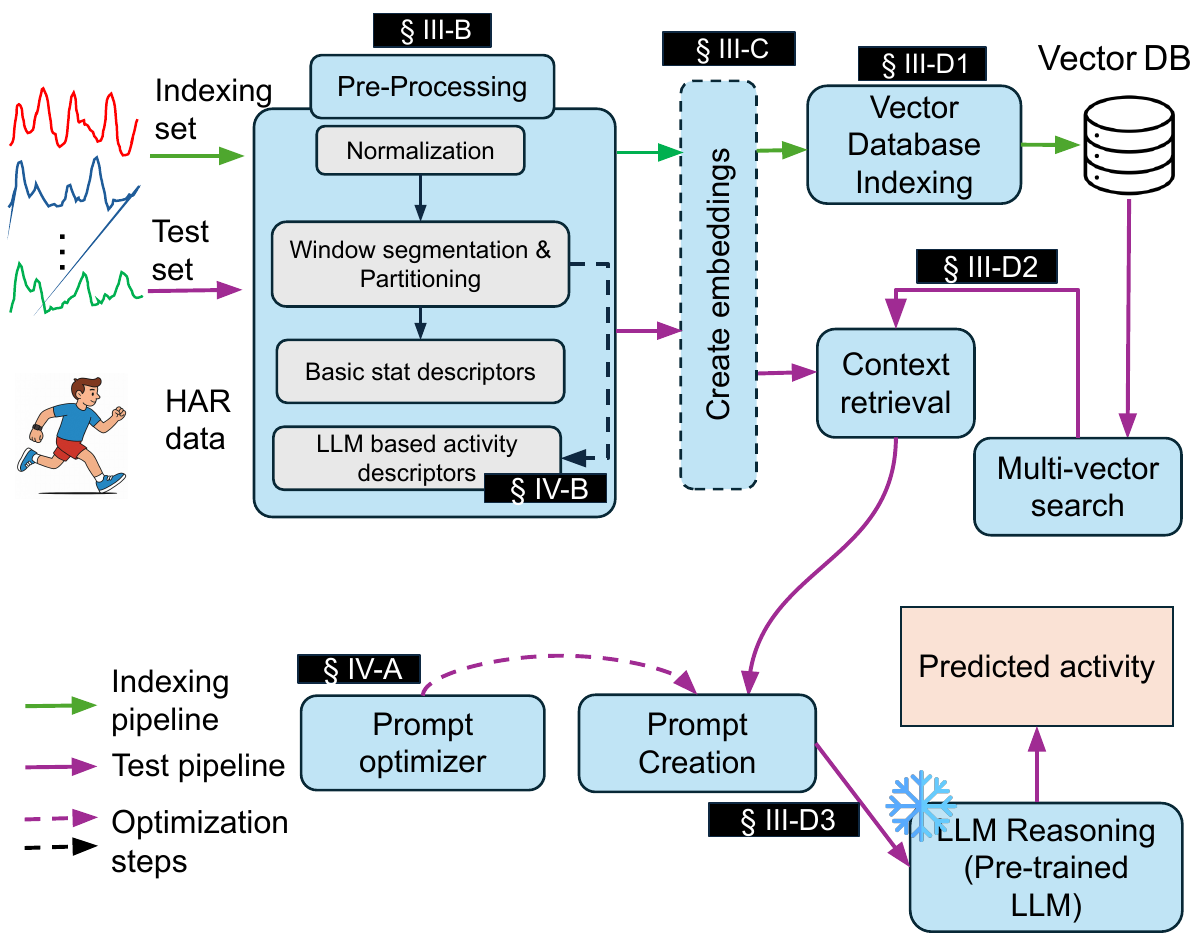}
        \caption{Overview of RAG-HAR framework}
        \label{fig:rag-har}
    \end{subfigure}
    \hfill
    \begin{subfigure}{.58\textwidth}
        \centering
        \includegraphics[width=\linewidth]{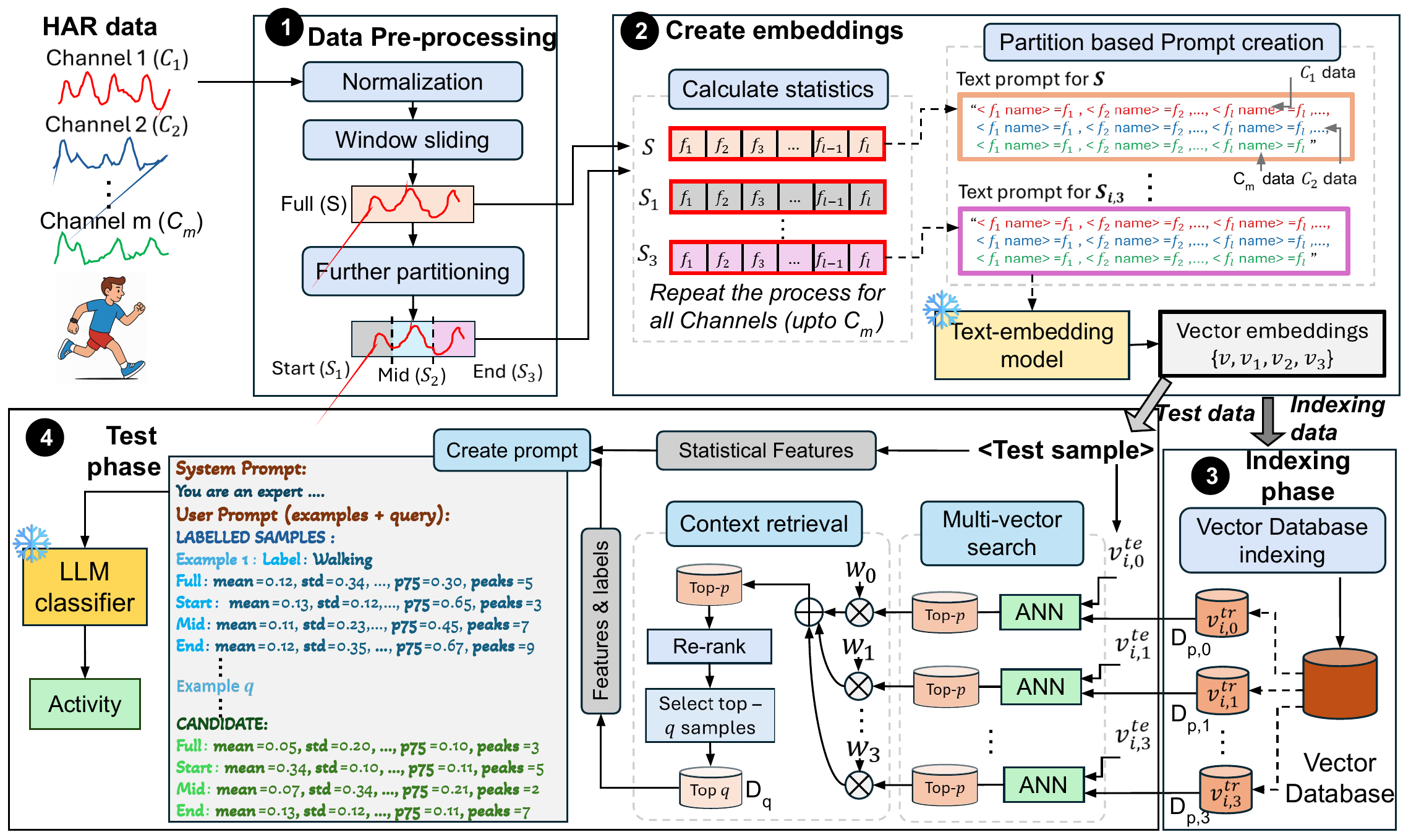}
        \caption{\solution{} Baseline stage}
        \label{fig:data2embedding}
    \end{subfigure}       
\caption{\solution{} architecture. Section numbers (§) point to where each step is detailed.}
\label{fig:enhanced-rag-har-strategies}
\end{figure*}

\section{Methodology}

\subsection{\solution{} overview}

As illustrated in Fig.~\ref{fig:rag-har}, the proposed two-stage \solution{} framework achieves training-free HAR by integrating semantic retrieval with LLM reasoning. In the \textit{Baseline} stage, raw activity signals are first passed through \textit{Data Processing} (\S~\ref{subsec:data-processing}), which involves normalization, window-based time-series slicing, partitioning, and extraction of statistical descriptors. These descriptors are then converted into embeddings (\S~\ref{subsec:create-embeddings}), which serve as the foundation for LLM-based HAR (\S~\ref{subsec:har-prediction}) under two pipelines (or phases). In the indexing pipeline, analogous to model training in traditional DL-based methods, embeddings from the indexing data are used to build a vector database (\S~\ref{subsubsec:indexing}). In the test pipeline (\S~\ref{subsubsec:inference-phase}), \solution{} retrieves contextual knowledge from this database to support accurate LLM predictions through structured prompts (\S~\ref{subsubsec:prompt-design}). The \textit{Optimized} stage of \solution{}~(\S~\ref{sec:optional-optimization}) enhance performance by refining prompt design for LLM prediction and incorporating natural language descriptors in place of statistical descriptors when creating embeddings. 




\subsection{Data processing}\label{subsec:data-processing}

\subsubsection{HAR data}
Typically contains  various types of sensor readings that capture human activity patterns and movements as time-series. These datasets are collected using off-the-shelf wearable devices (e.g., smartwatches, smart shoes) as well as smartphones. As illustrated in Fig.~\ref{fig:data2embedding}, HAR data may consist up to {$m$} channels  representing either different sensor modalities (e.g., accelerometer, gyroscope) or dimensions (e.g., $X$, $Y$, and $Z$ axes of accelerometer data) or body parts.

\subsubsection{Data pre-processing}
\textbf{Existing work and limitations:}
This is an essential step in HAR to convert raw time-series signals into structured representations making them suitable for downstream analysis. Many prior works~\cite{khaertdinov2021deep, suh2022adversarial} leveraged mechanisms, which segment time-series into small, fixed-size overlapping windows and extract features considering the entire window.  
However, computing statistical features over the entire window can suppress fine-grained temporal cues, such as transient events or non-periodic localized patterns that occur within small part of the window, leading to the loss of important features critical for HAR.


\noindent
\textbf{Our approach:} We retain a sliding-window framework for fair comparison with benchmarks, but each window is subdivided into further partitions, and statistics are computed for the entire sliding-window and each partition separately. This mechanism preserves both short-lived transients and longer-horizon trends, yielding a feature space that is measurably richer than single-window summaries.

As shown in Fig.~\ref{fig:data2embedding}-1, there are three main steps in data pre-processing. 
\textit{i})~\textbf{Normalization:} All raw sensor channels, $C$, are standardized using Z-score normalization. This normalization step converts all channels to a common scale, thereby reducing the potential impact of large values corrupting the internal attention mechanism.
\textit{ii})~\textbf{Sliding window-based segmentation:} Following the protocols used in prior HAR studies \cite{khaertdinov2021deep, suh2022adversarial, ahmad2023alae, haresamudram2023investigating}, the time-series distribution is partitioned into segments with a fixed-size window $L$ and step size $D$. We denote the segment as $S$. There are two major advantages of this process. First, it captures short-term temporal context, which is important for HAR. Second, it produces fixed-size units for the ease of computing in later steps of the \solution{} pipeline.
\textit{iii})~\textbf{Further partitioning:} For a given segment $S$, we extract statistical features (see \S~\ref{subsubsec:stat extract}) considering the {\textit{Full}} segment, $S$ and three equally partitioned sub-segments (\textit{Start}, \textit{Mid}, and \textit{End}), denoted by $S_{k}$, where $k\in \{1,2,3\}$.
Empirically, we observed that these four segments (i.e., $S$, $S_{1}, S_{2}, \text{and}\ S_{3}$)  can effectively identify key feature variations of HAR traces both globally (i.e., behavioral trends) and locally (i.e., transient events). For example, activities such as standing and sitting show fairly constant fluctuations throughout the entire segment, and activities such as cycling and jumping show high dynamics at the beginning (i.e., in $S_{1}$) compared to other sub-segments.

\subsubsection{Statistical value calculation}\label{subsubsec:stat extract}
For $S$ and each $S_{k}$, statistical features ($f$) (up to $l$ features) are computed across all channels  as denoted in Fig.~\ref{fig:data2embedding}-2. Based on our analysis, we consider mean, max, min, $25^{th}$ percentile (Q1), $75^{th}$ percentile (Q3), standard deviation, median, and no. of peaks of the time-series distribution by setting $l=8$.  These basic descriptors capture importat behavioral patterns including central tendency, variability, distributional shape, and frequency of noticeable events of an activity.  By calculating feature values for $S$ and each $S_{k}$, we ensure that statistical features are captured at both global and local levels respectively.   The $l$ number of features computed for each of the channel $C$  in either $S$ or a given $S_{k}$ are arranged into a single feature vector of $l \cdot m$ elements. Each $S$ yields four such vectors (three vectors from each $S_{k}$ and $S$ itself), which collectively form the basis for multi-vector retrieval in LLM-based activity prediction.

\subsection{Create embeddings} \label{subsec:create-embeddings}



\textbf{Existing work and limitations:} Prior work \cite{ahmad2023alae, haresamudram2023investigating} typically requires dataset-specific model training, often coupled with tailored data augmentation and careful optimization for each dataset introducing large computational overheads.

\noindent
\textbf{Our approach:} Since no general-purpose time-series embedding models currently exist, we avoid dataset-specific training by using a general-purpose text embedding model pretrained on large-scale public datasets. Because such models are optimized for natural language rather than decimal-heavy numerical sequences, directly embedding raw time series is ineffective. Instead, our approach converts feature vectors of $S$ and each $S_{k}$ into structured text strings using a fixed template, yielding compact, interpretable embeddings with clear semantic meaning.

There are two main steps in this process: \textit{i})~\textbf{Serialization to text:} The feature vectors derived for $S$ and each $S_{k}$ (see \S~\ref{subsubsec:stat extract}) is converted into a short textual string using a fixed template, e.g., $\langle f_l\ name \rangle = f_l$, as illustrated in Fig.~\ref{fig:data2embedding}-2. Here, $\langle f_l\ name \rangle$ is replaced by the feature name (e.g., mean, maximum, etc.), and $f_l$ denotes the corresponding feature value. Since a given vector contains feature values from all channels, the resulting string for $S$ and each $S_{k}$ provides a consistent representation of feature names and their associated values across all $C$. Text serialization offers three benefits.  First, since the variable component of the  string  correspond solely  to  feature values, the resulting embeddings can effectively capture the feature variations. Second, the serialized format remains compact enough to fit within the LLM context size. Third, by incorporating feature names alongside their values, it provides natural language context that LLMs can readily interpret.
\textit{ii})~\textbf{Generate embeddings:}
Strings corresponding to $S$ and each $S_{k}$ are encoded into vectors ($\bm{v}\ \text{and}\ \bm{v_{k}} \in\mathbb{R}^d$) separately. We use a publicly available pre-trained embedding model to generate embeddings. Using a pre-trained model eliminates the need for training an embedding model on HAR-specific data, and since it is trained on a large and diverse text corpus (including numeric tokens), the resulting embeddings effectively capture variations in feature values.


\subsection{HAR activity prediction}\label{subsec:har-prediction}

\textbf{Exisitng work and limitations.} Existing HAR pipelines \cite{ahmad2023alae, suh2022adversarial} typically rely on supervised DL models trained end-to-end on labeled data. These approaches require large annotated datasets, costly training cycles, and extensive hyperparameter tuning. 

\noindent
\textbf{Our approach.} In contrast, \solution{}  does not involve conventional model training, instead generates embeddings and creates a comprehensive vector database that will be used during the Test phase with pre-trained LLMs.  Since we employ pre-trained LLMs, this approach is training-free, computationally lightweight, and adaptable across datasets, as the LLM has already been trained on large corpora of data.

\subsubsection{Vector Database Indexing}\label{subsubsec:indexing}
The first phase of HAR activity prediction is to build a database containing embeddings from the training dataset (Fig.~\ref{fig:data2embedding}-3). Following the approach in \S~\ref{subsec:data-processing}, we generate embeddings from the training data. For the $i$-th segment from training set denoted by $S_i$ 
(i.e., $i\in \mathcal{I}^{tr}$ - training sample indices) 
four embeddings 
$V_i^{tr} = \{\bm{v}_{i,k}^{tr} | k\in \{0,1,2,3\}\}$,
are created (for $S_i$ and each partition $S_{i,k}$) and stored in the vector database along with metadata including the segment id ($i$), sub-segment id ($k$), activity label, user id, and statistical features prior to embedding. Here, $\bm{v}_{i,0}^{tr}$ corresponds to the vector embedding from $S_i$.

\subsubsection{Test/Inference phase}
\label{subsubsec:inference-phase}
Similar to indexing phase, we first create the four vector embeddings 
$V_j^{te}= \{\bm{v}_{j,k}^{te} | j\in \{0,1,2,3\}\}$
for the $j$-th sample ($S_j$) from the test data ($j\in \mathcal{I}^{te}$). Next, as shown in Fig.~\ref{fig:data2embedding}-4, we conduct context retrieval to select top-$q$ samples from the vector database that closely match with the given test sample prior to the LLM-based prediction. As the first step, for each \(\bm{v}_{j,k}^{te}\in V_j^{te}\) we run an Approximate Nearest Neighbor (ANN) search using cosine similarity, comparing \(\bm{v}_{j,k}^{te}\) against the corresponding training embeddings: $\{\bm{v}_{i,k}^{tr} \mid i \in \mathcal{I}^{tr}\}$.
This retrieval returns, for each \(\bm{v}_{j,k}^{te}\), a ranked list of the top-\(p\) nearest training embeddings among \(\{\bm{v}_{i,k}^{tr} \mid i \in \mathcal{I}^{tr}\}\) along with their ANN scores, which we denote as \(D_{p,k}\). In the second step, we perform weighted re-ranking by summing the ANN scores from the selected vector components in each \(D_{p,k}\), weighted by \(w_k\), where \(w_k \geq 0\) and \(\sum_k w_k = 1\). This weighting mechanism prioritizes candidates according to the relative importance of the corresponding vector embeddings, in other words, different sub-segments. Based on the updated ANN scores, we re-rank the union of the \(D_{p,k}\) sets and select the final top-\(q\) vector samples \(\{\bm{v}_{i,k}^{tr} \mid i \in \mathcal{I}^{q}, k \in \{0,1,2,3\}\}\). We denote this context set by $D_q$.

\subsubsection{Prompt design and LLM prediction}\label{subsubsec:prompt-design}
In the prompt design step, given the selected vector embeddings in $D_q$ and a test segment $S_j$, we construct a text prompt as shown in  Fig.~\ref{fig:data2embedding}-4, The prompt has two main components: the \textit{system} prompt, which instructs the LLM about the classification task, and the \textit{user} prompt, which provides the data required for activity prediction.  The \textit{Labeled} samples section conveys information from the training set $D_q$.  Each labeled sample in the prompt includes the activity label and the corresponding statistical features of $S_i$ and $S_{i,k}$. The \textit{Candidate} section contains only the statistical features extracted from the test sub-segments $S_j$ and $S_{j,k}$. 

During LLM-based prediction, the model processes this prompt and outputs the activity label (along with a brief rationale).  The prediction is grounded not only in the LLM’s pretrained knowledge but also in the statistical features retrieved from $D_q$, thereby leveraging both prior knowledge and context from the training data.






\begin{figure}
    \centering
    \includegraphics[width=0.90\linewidth]{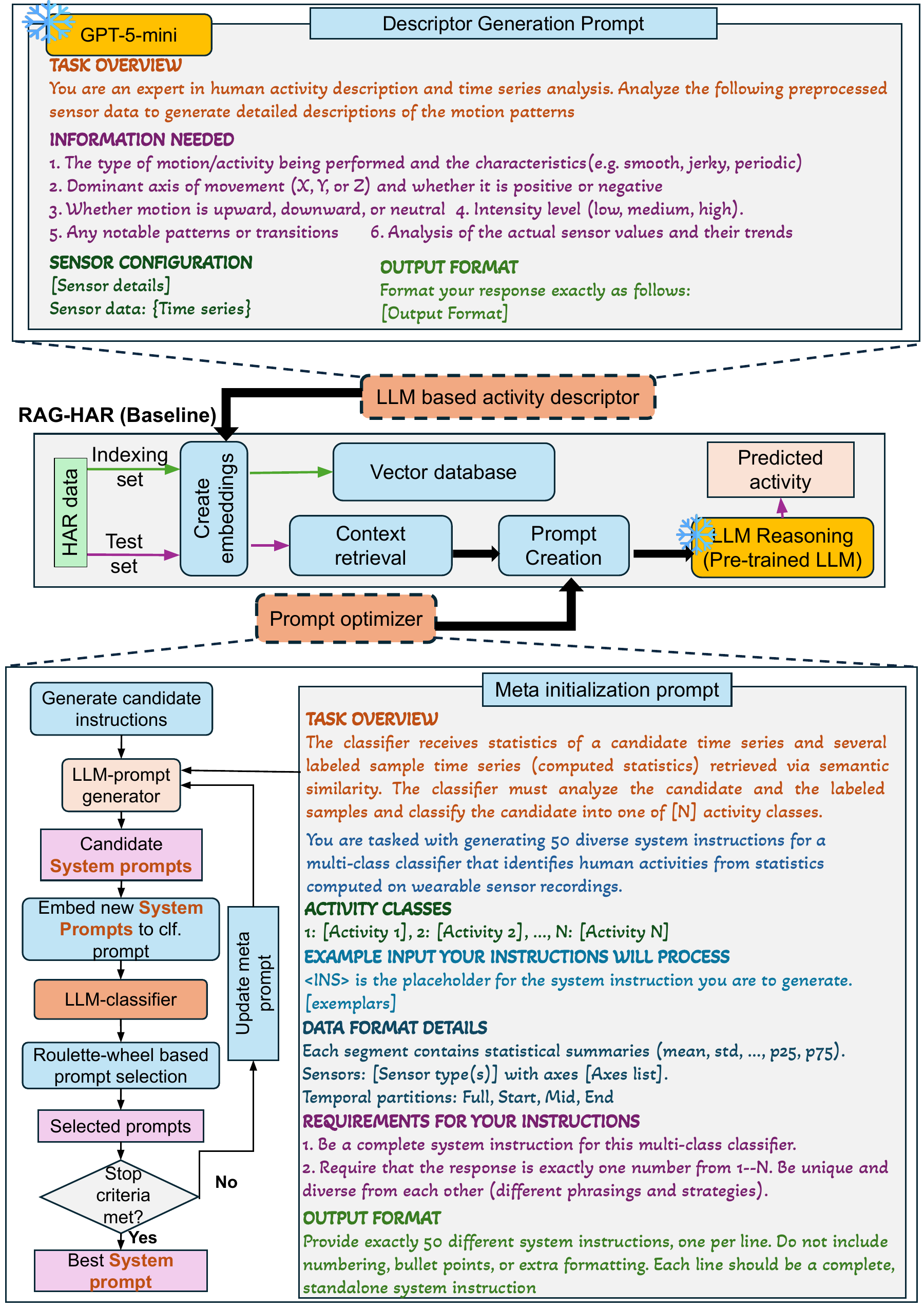}
    \caption{Further optimization techniques for \solution{}.}
    \label{fig:optimizers}
\end{figure}

\section{Further optimization steps}
\label{sec:optional-optimization}
Building on the \textit{Baseline} stage, we introduce the \textit{Optimization} stage of \solution{} integrating the following steps.

\subsection{Prompt optimization}
\label{subsec:promptopt}

Optimizing the \textit{system} prompt that serve as the instruction to the LLM would enhance \solution{}. In general, LLM performance is highly sensitive to instruction wording, example ordering, and the way evidence is presented. Therefore, a carefully optimized prompt can substantially improve classification accuracy and reduce spurious outputs and is conceptually equivalent to hyperparameter tuning of an already-trained model~\cite{fernando2023promptbreeder}.
Thus, inspired by~\cite{fernando2023promptbreeder, yang2023large, guo2025evoprompt}, we design our prompt optimizer framework as shown in Fig. \ref{fig:optimizers}.
In this process, the LLM classifier remains fixed with no weight updates, since the method does not involve training but instead operates as a meta-level optimization over prompts.
Below we have described the key-steps of this process.

\noindent
\textbf{Step 1. Generate $M$ candidate instructions:}  
We start by generating a diverse pool of $M$ initial instructions, referred to as candidate instructions. These candidates are automatically produced by optimizer model using the meta-initialization prompt illustrated in Fig.~\ref{fig:optimizers}. In addition to the meta prompt, we provide exemplars that include a few \textit{Labeled} and \textit{Candidate} samples, along with their statistical features, to help the model understand the \textit{user} prompt the LLM-classifier will receive. This comprehensive prompt design enhances the diversity of the generated instructions, which is essential for covering a wide region of the prompt space and avoiding premature convergence to suboptimal phrasings.\\
\noindent
\textbf{Step 2. Evaluate prompts using LLM-classifier:}
Each candidate instruction is treated as the \textit{system} prompt in the main input to the LLM-classifier to generate predictions for the validation set. We compute F1-Score as the fitness score for the given instruction.\\  
\noindent
\textbf{Step 3. Roulette-wheel based sampling:}  
From the evaluated pool of candidate instructions, we select a subset of $r$ instructions using roulette-wheel selection. The chosen $r$ instructions are then used in the next iteration of instruction generation, as outlined in \textbf{Step 4}. Each candidate’s probability of selection is proportional to its validation fitness. This strategy balances exploitation, by favoring high-performing instructions, with exploration, by allowing lower-performing templates the opportunity to contribute\cite{guo2025evoprompt}.\\
\noindent
\textbf{Step 4. Meta-prompt construction:}  
The selected $r$ instructions are embedded into the meta-prompt provided to the LLM for instruction generation in \textbf{Step 1}. The modified prompt includes, for each selected instruction, its text, validation fitness, and a small set of exemplars as we provided in \textbf{Step 1}. This contextualization enables the optimizer to infer which instruction characteristics are associated with higher fitness.\\
\noindent
\textbf{Step 5. Generate instructions iteratively:}  
Using the modified meta-prompt, the LLM generates $r$ number of new instructions. These instructions then undergo the same process from \textbf{Step 2} to \textbf{Step 4}, repeatedly generating instructions for up to $P$ iterations or until a stopping condition is reached (no improvement for $T$ consecutive iterations). This iterative procedure allows the optimizer to progressively refine prompts by combining the strengths of prior high-performing instructions, while increasing search throughput and keeping per-iteration cost and latency manageable. Throughout the process, we track the best-performing instruction and maintain a versioned log to ensure reproducibility. After termination, the single best instruction is selected and used as the system prompt for the classifier on the held-out test set.




\noindent
\textbf{Prompt optimization strategies:}  
We consider 3 strategies for prompt optimization to improve the quality of generated prompts.
\textit{i})~\textbf{Exploration:}~In the initial iterations, the model is prompted to generate a diverse set of instructions. This encourages the optimizer to explore a broad range of possibilities.
\textit{ii})~\textbf{Combination:}~In subsequent iterations, the model applies crossover and mutation to the best-performing instructions. This enables the optimizer to create new candidates by combining key elements from the best-performing instructions.
\textit{iii})~\textbf{Refinement:}~In the final stage, the model makes literal adjustments to the instructions, rephrasing them to test which wordings elicit the best performance from the classifier.

\subsection{LLM-Based Descriptors for Enhanced Retrieval}

In \S~\ref{subsubsec:inference-phase}, we detailed the context retrieval process and a more informative retrieval process is always advantageous for \solution{}, as it can lead to higher accuracy and reduce the number of context retrievals. To utilize the text-embedding model for vectorization without requiring additional training, we initially generated simple templates with basic statistics programmatically for dataset vectorization (see \S~\ref{subsec:create-embeddings}). However, such templates capture shallow statistical features. A more effective approach is to construct detailed natural language descriptors that highlight trends in the time series, capture sensor configuration details, and incorporate comprehensive statistical analysis for each activity window.

To achieve this level of descriptive fidelity, we designed a structured prompt for the LLM. The prompt provides the raw time-series data alongside sensor configuration details (sampling rate, sensor placement, orientation, and axis mappings) and instructs the model to analyze the motion characteristics comprehensively. The model output follows a structured format (full-segment and segment-wise analyses), ensuring both consistency and coverage of critical aspects such as dominant axis, motion smoothness, intensity, and transitions.
The rest of the \solution{} remains unchanged: once generated, these natural language descriptors are vectorized using the text-embedding model and indexed in the vector database. At inference time, they enhance retrieval by providing richer, semantically meaningful representations of the activities. 

\section{Experiment setup}


\subsection{Datasets}\label{subsec:dataset}

We evaluate \solution{} on six widely used HAR benchmarks—HHAR~\cite{stisen2015smart}, PAMAP2~\cite{reiss2012introducing}, MHEALTH~\cite{banos2014mhealthdroid}, GOTOV~\cite{paraschiakos2020activity}, SKODA~\cite{stiefmeier2008wearable}, and USC-HAD~\cite{zhang2012usc}—which span 6 to 60 sensor channels and 6 to 16 activities, covering diverse activity types such as locomotion (e.g., walking, running), daily living tasks (e.g., ironing, vacuuming), fitness exercises (e.g., cycling, jogging), manufacturing assembly-line tasks, and settings that capture activity variations across pace, direction, or equipment.

\subsubsection{HHAR~\cite{stisen2015smart}}
This dataset primarily contains locomotion-style activities such as walking, sitting, standing, and going up/down the stairs. The dataset consists of sensor data collected from smart phones and smart watches. Following ~\cite{haresamudram2023investigating}, we use only sensor data at wrist part which consists of 6 channels. The window size is set to 2 seconds, with an overlap of 50\%. Data preparation for indexing and testing follows ~\cite{haresamudram2023investigating}. 

\subsubsection{PAMAP2~\cite{reiss2012introducing}}
The dataset originally contains 52 channels, including heart rate and full IMU sensor streams. Following \cite{suh2022adversarial}, for this work, we retained 36 channels by excluding heart rate, temperature, and orientation data, as orientation was reported invalid during collection. We focused on 12 protocol-defined activities recorded from 8 subjects. In addition to locomotion activities, PAMAP2 contains daily living-room activities such as ironing and vacuum cleaning. The IMU data was sampled at 100 Hz, and we applied a sliding window of 200 samples (2 seconds) with a step size of 50 samples (0.5 seconds).  Data preparation for indexing and testing follows ~\cite{suh2022adversarial}.

\subsubsection{MHEALTH~\cite{banos2014mhealthdroid}}
The dataset was sampled at 50 Hz and contains 21 channels. In addition to locomotion activities, it also includes exercises such as jogging, cycling and etc. This has 12 activities in total. Following \cite{suh2022adversarial}, we use sliding windows of 200 samples (4 seconds) with a step size of 50 samples (1 second). Data preparation for indexing and testing follows ~\cite{suh2022adversarial}.

\subsubsection{GOTOV~\cite{paraschiakos2020activity}}
The dataset contains 9 sensor channels collected from three body positions on 35 older adults aged over 61 while performing 16 activities. This dataset contains same activities in different paces (e.g.  walking slow, walking normal and walking fast) and same acitivity with different equipments (e.g. sitting on sofa, sitting on chair and sitting on couch).  
Six participants were excluded due to missing sensor data. Following ~\cite{ahmad2023alae}, windows were created using 24-sample segments with 50\% overlap between consecutive windows. Data preparation for indexing and testing follows ~\cite{ahmad2023alae}.

\subsubsection{SKODA~\cite{stiefmeier2008wearable}}
This dataset contains recordings of 10 assembly-line activities performed by car manufacturing workers, captured through 60 sensor channels positioned on the right-hand side of the body. Following ~\cite{ahmad2023alae}, windows were created using 24-sample segments with 50\% overlap between consecutive windows. Data preparation for indexing and testing follows ~\cite{ahmad2023alae}.

\subsubsection{USC-HAD~\cite{zhang2012usc}}
This dataset includes accelerometer and gyroscope signals, each with three degrees of freedom, resulting in six channels per time step. Data was collected from 14 subjects performing 12 basic activities such as walking in different directions, running, and jumping. Following ~\cite{khaertdinov2021deep}, the dataset was downsampled to 33.3 Hz, segmented into 1-second windows with 50\% overlap, and data from subjects 13 and 14 was reserved for testing. Data preparation for indexing and testing follows ~\cite{khaertdinov2021deep}.

\subsection{Evaluation setup}\label{subsec:evaluation-setup}

We generated embeddings using \textbf{text-embedding-3-small}~\cite{openaiIntroducingEmbedding}, as this model consistently provided more stable retrieval performance compared to alternatives. These embeddings (dimension = 1536) were indexed in \textbf{Zilliz}~\cite{zilliz}, a high-performance cloud vector database, which natively supports the embedding output size. During retrieval, we applied \textbf{weighted re-ranking} with weights (0.4, 0.2, 0.2, 0.2). Among the segmentations tested, the [full, start, mid, end] configuration yielded the best balance of coverage and accuracy. We fixed the number of \textbf{retrieved contexts} ($q$) to 10, as higher values increased inference cost without accuracy gains, while lower values reduced robustness. For the classification stage, we employed \textbf{gpt-5-mini}~\cite{openaiIntroducingGpt5}, which offered the best trade-off between accuracy and cost after experiments with other models (openai, gemini and llama variants). Finally, for prompt optimization, we used \textbf{gpt-5}~\cite{openaiIntroducingGpt5}, leveraging its stronger reasoning for more effective refinements.

\section{Results}

\subsection{Benchmarking baseline \solution{}}

To assess the performance, \solution{} is compared with SOTA HAR models from prior \textbf{PerCom} publications. Table~\ref{tab:results-comparison-table} shows that baseline \solution{} consistently outperforms the SOTA across multiple HAR datasets. On PAMAP2, Skoda, and GOTOV, \solution{} delivers gains of about 1–2\% in F1-score over the strongest prior models, while on MHEALTH \solution{} achieves a marginal yet consistent improvement over the best adversarial feature extractors. For more challenging datasets such as HHAR, \solution{} improves performance by over 7\% F1-Score compared to SimCLR and DeepConvLSTM baselines and remains competitive with the SOTA methods. On USC-HAD, although slightly below the triplet-LSTM variant, \solution{} still outperforms conventional CNN, LSTM, and Transformer-like baselines. To further strengthen the performance on this dataset, we evaluate the optimized stage of \solution{} in \S~\ref{subsec:dissection}. 

Although the performance gains may appear modest in some datasets, they are particularly noteworthy for two reasons. First, \solution{} achieves these improvements without any pretraining, thereby reducing the additional time and resources often required by self-supervised or adversarial approaches. Second, by outperforming diverse baselines across multiple datasets, \solution{} demonstrates a level of robustness and generalization that is difficult to achieve with other benchmark models without extensive task-specific retraining. Overall, these results highlight \solution{}’s strong adaptability across different sensor environments, setting new benchmarks in five out of six datasets.

\begin{table}[!t] 
\centering
\scriptsize
\caption{Benchmarking \textbf{Baseline} \solution{} with SOTAs. The reference shows from where the values were taken.}
\label{tab:results-comparison-table}
\renewcommand{\arraystretch}{1.2}
\begin{tabular}{|l|l|l|l|}
\hline
\textit{\textbf{Dataset}} & \textit{\textbf{Model}}                         & \textit{\textbf{Acc(\%)}} & \textit{\textbf{F1}(\%)}   \\ \hline
\multirow{10}{*}{PAMAP2}   & MC-CNN ~\cite{suh2022adversarial}            & 79.77               & 72.72 \\ \cline{2-4} 
                          & DeepConvLSTM ~\cite{suh2022adversarial}       & 76.23              & 67.54 \\ \cline{2-4} 
                          & Transformer-like ~\cite{suh2022adversarial}    & 82.88              & 74.84 \\ \cline{2-4} 
                          & METIER    ~\cite{suh2022adversarial}         & 83.97               & 77.66 \\ \cline{2-4} 
                          & ADFE ~\cite{suh2022adversarial}  [Percom '22]    & 85.69              & 77.84  \\ \cline{2-4} 
                          & Softmax LSTM ~\cite{khaertdinov2021deep}              & -                 & 79.20      \\ \cline{2-4} 
                          & Triplet LSTM ~\cite{khaertdinov2021deep}            & -                          & 81.10      \\ \cline{2-4} 
                          & b-LSTM-S ~\cite{khaertdinov2021deep}             & -                          & 86.80          \\ \cline{2-4} 
                          & Hargpt ~\cite{ji2024hargpt}                     &  32.11    & 31.57  \\ \cline{2-4}
                          & LLM as Virtual Annotators ~\cite{hota2025evaluating}                     &  56.70    & 53.80  \\ \cline{2-4}
                          & Triplet LSTM (OTL) ~\cite{khaertdinov2021deep}[Percom '21]   & -                  & \underline{90.40}    \\ \cline{2-4} 
                          & \textbf{\solution{}  (ours)}                      & \textbf{91.60}                & \textbf{91.12}      \\ \hline
\multirow{8}{*}{MHEALTH} & MC-CNN  ~\cite{suh2022adversarial}           & 89.69               & 87.42  \\ \cline{2-4} 
                          & DeepConvLSTM ~\cite{suh2022adversarial}   & 89.24              & 87.17 \\ \cline{2-4} 
                          & Transformer-like ~\cite{suh2022adversarial}    & 87.44    & 85.13  \\ \cline{2-4} 
                          & METIER  ~\cite{suh2022adversarial}      & 94.42              & 94.09  \\ \cline{2-4} 
                          & ADFE ~\cite{suh2022adversarial}    [Percom '22]        & \underline{96.72}         & \underline{96.47}  \\ \cline{2-4} 
                          & Softmax LSTM ~\cite{khaertdinov2021deep}                       & -                          & 44.20      \\ \cline{2-4}  
                          & Triplet LSTM (HTL-SB) ~\cite{khaertdinov2021deep} [Percom '21]  & -                & 65.60              \\ \cline{2-4}  
                          & \textbf{\solution{}  (ours)}         & \textbf{96.91}              & \textbf{96.74}         \\ \hline
\multirow{7}{*}{GOTOV}    & LSTM Learner Baseline  ~\cite{ahmad2023alae}         & -                          & 61.10            \\ \cline{2-4} 
                          & DeepConvLSTM ~\cite{ahmad2023alae}                   & -                          & 66.90              \\ \cline{2-4} 
                          & b-LSTM-S ~\cite{ahmad2023alae}                & -                          & 63.90         \\ \cline{2-4} 
                          & Att. Model ~\cite{ahmad2023alae}             & -                          & 70.70       \\ \cline{2-4} 
                          & Attend and Discriminate ~\cite{ahmad2023alae}     & -                          & 76.20             \\ \cline{2-4} 
                          & ALAE-TAE-CutMix+  ~\cite{ahmad2023alae} [Percom '23]      & -                          & \underline{79.40}  \\ \cline{2-4} 
                          & \textbf{\solution{} (ours)}                        & \textbf{81.55}                & \textbf{79.92}                         \\ \hline
\multirow{7}{*}{Skoda}    & LSTM Learner Baseline ~\cite{ahmad2023alae}  & -                          & 90.40        \\ \cline{2-4} 
                          & DeepConvLSTM ~\cite{ahmad2023alae}   & -                          & 91.20      \\ \cline{2-4} 
                          & b-LSTM-S  ~\cite{ahmad2023alae}      & -                          & 92.10             \\ \cline{2-4} 
                          & Att. Model~\cite{ahmad2023alae}         & -                          & 91.30            \\ \cline{2-4} 
                          & Attend and Discriminate  ~\cite{ahmad2023alae}       & -                          & 92.80              \\ \cline{2-4} 
                          & ALAE-TAE-CutMix+  ~\cite{ahmad2023alae} [Percom '23]   & -             & \underline{94.80}   \\ \cline{2-4} 
                          & \textbf{\solution{} (ours)}     & \textbf{96.04}                & \textbf{95.21}       \\ \hline
\multirow{6}{*}{USC-HAD} & Softmax LSTM baseline ~\cite{khaertdinov2021deep}       & -                          & 49.00    \\ \cline{2-4} 
                          & Triplet LSTM baseline ~\cite{khaertdinov2021deep}    & -                          & 53.50      \\ \cline{2-4} 
                          & DeepConvLSTM   ~\cite{khaertdinov2021deep}      & -                          & 46.00  \\ \cline{2-4} 
                          & Transformer-like architecture ~\cite{khaertdinov2021deep}         & -            & 55.00    \\ \cline{2-4}      
                          & Triplet LSTM (HTL-SB) ~\cite{khaertdinov2021deep} [Percom '21]     & -     & \underline{\textbf{62.80}}    \\ \cline{2-4} 
                          & \textbf{\solution{}  (ours)}                  & 57.20                & 58.63   \\ \hline
\multirow{6}{*}{HHAR}       & CPC ~\cite{haresamudram2023investigating}   & -                          & 59.17 \\ \cline{2-4} 
                          & GRU Classifier ~\cite{haresamudram2023investigating}        & -                          & 45.23     \\ \cline{2-4} 
                          & DeepConvLSTM ~\cite{haresamudram2023investigating}        & -                          & 52.37  \  \\ \cline{2-4} 
                          & SimCLR ~\cite{haresamudram2023investigating}   & -                & 52.84            \\ \cline{2-4}      
                          & Enhanced CPC ~\cite{haresamudram2023investigating}    [Percom '23]     & -           & \underline{59.25}   \\ \cline{2-4} 
                          & \textbf{\solution{}  (ours)}       & \textbf{58.61}                & \textbf{59.86}      \\ \hline
\end{tabular}
\end{table}



\subsection{Dissecting \solution{}: Baseline and Optimized Versions}
\label{subsec:dissection}

Fig.~\ref{fig:ablation-component} presents the F1-score (as the datasets are imbalanced) achieved by different components of \solution{} on the MHEALTH and USC-HAD datasets (best-performing and least-performing datasets respectively).

\subsubsection{Baseline \solution{}}

\begin{figure*}[t]
    \centering
    \begin{subfigure}{.29\textwidth}
        \centering
        \includegraphics[width=\linewidth]{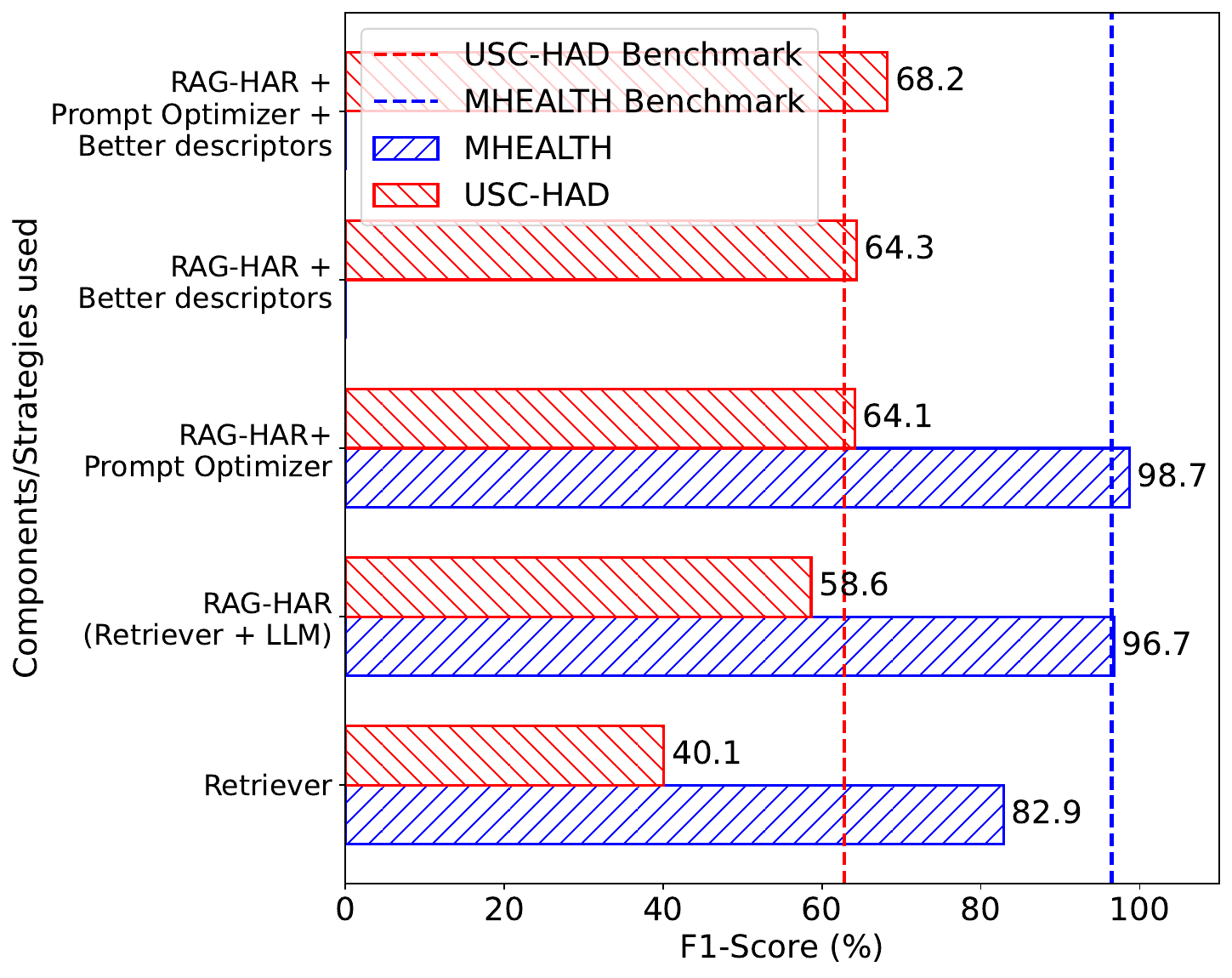}
        \caption{\textbf{Strategies enhancing the \solution{}}}
        \label{fig:ablation-component}
    \end{subfigure}
    \hfill
    \begin{subfigure}{.345\textwidth}
        \centering
        \includegraphics[width=\linewidth]{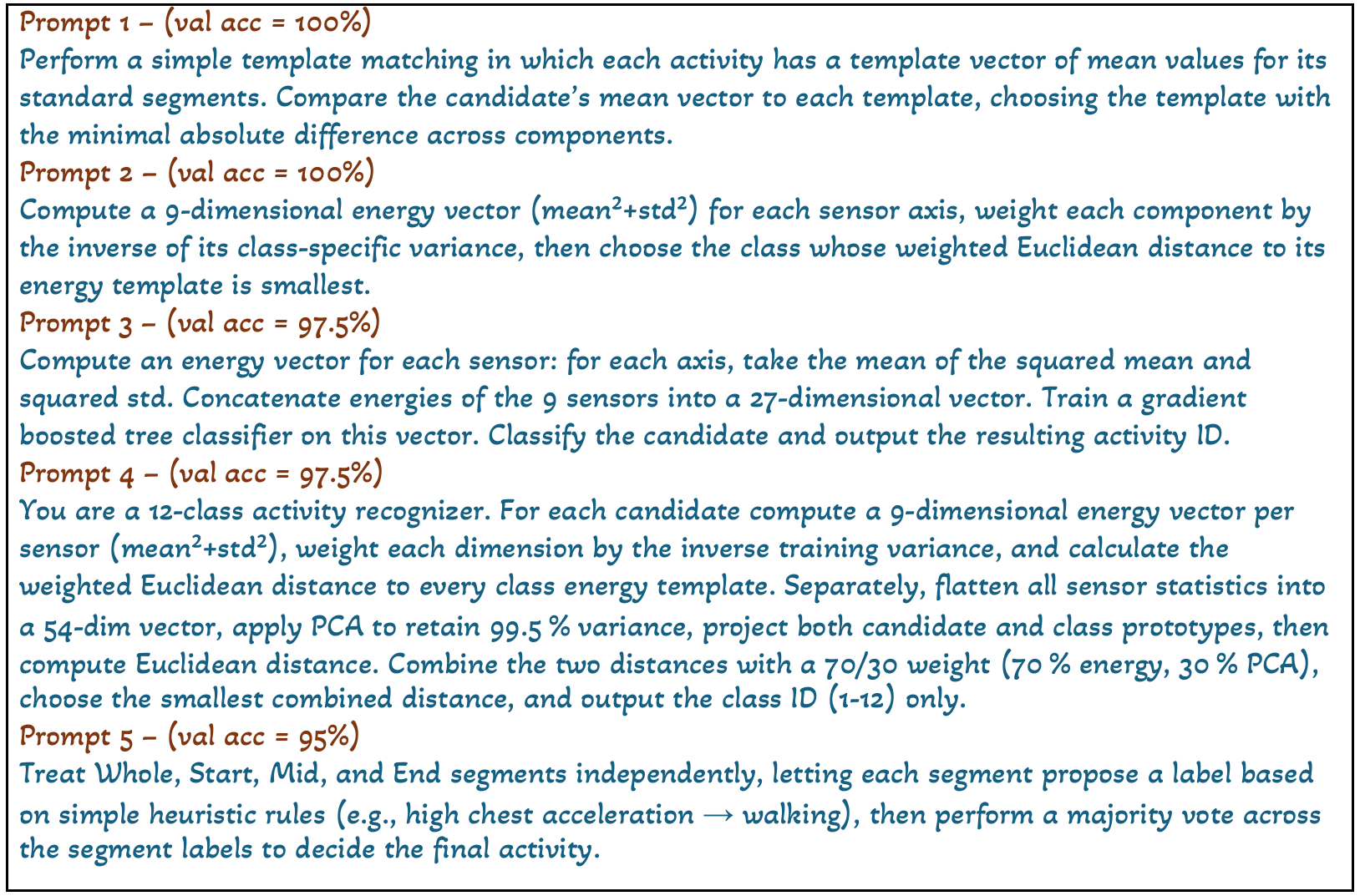}
        \caption{\textbf{Optimized prompts (MHEALTH)}}
        \label{fig:mhealth-opt-prompts}
    \end{subfigure}    
    \hfill
    \begin{subfigure}{.345\textwidth}
        \centering
        \includegraphics[width=\linewidth]{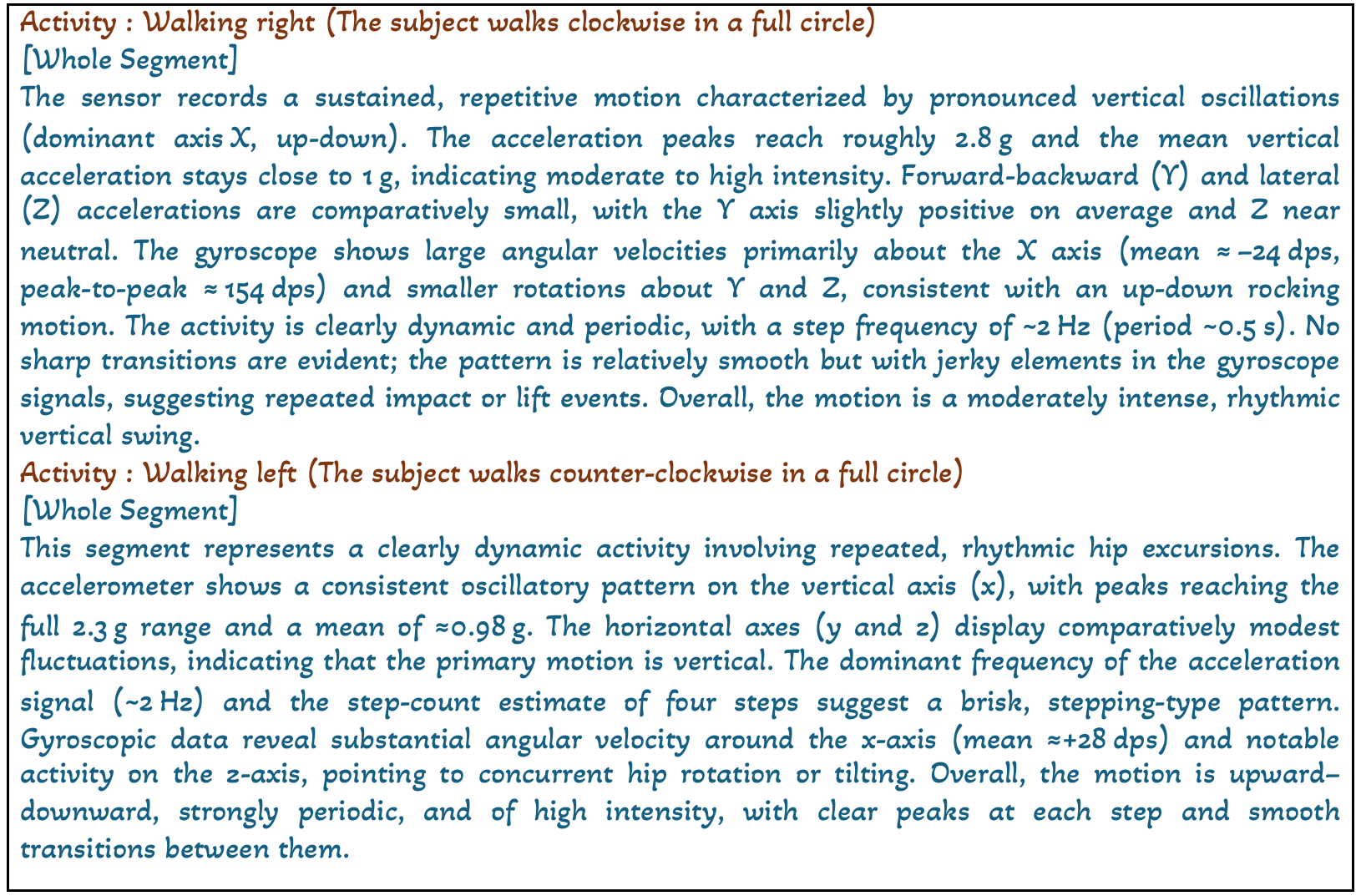}
        \caption{\textbf{Activity descriptor (USC-HAD)}}
        \label{fig:activity-descriptor}
    \end{subfigure}    
\caption{Enhanced \solution{} strategies}
\label{fig:enhanced-rag-har-strategies}
\end{figure*}

We first analyze the performance of the retriever using the embeddings stored in the vector database. Predictions are made by constructing a distribution of cosine similarity scores, applying a threshold of 0.75, and assigning the majority class (i.e., the most frequent label within the retrieved set) as the output. Fig.~\ref{fig:ablation-component} shows that F1-Score of retriever is around 40.1\% and 82.9\% for the USC-HAD and MHEALTH datasets respectively.
While this provides a reasonable baseline, performance is still low as it struggles in cases where activities share highly similar motion patterns across subjects. For example, the retriever often confuses \textit{running} and \textit{jogging}, since the embedding model captures only minimal differences between these activities.

\begin{figure}[ht]
    \centering
    \includegraphics[width=\columnwidth]{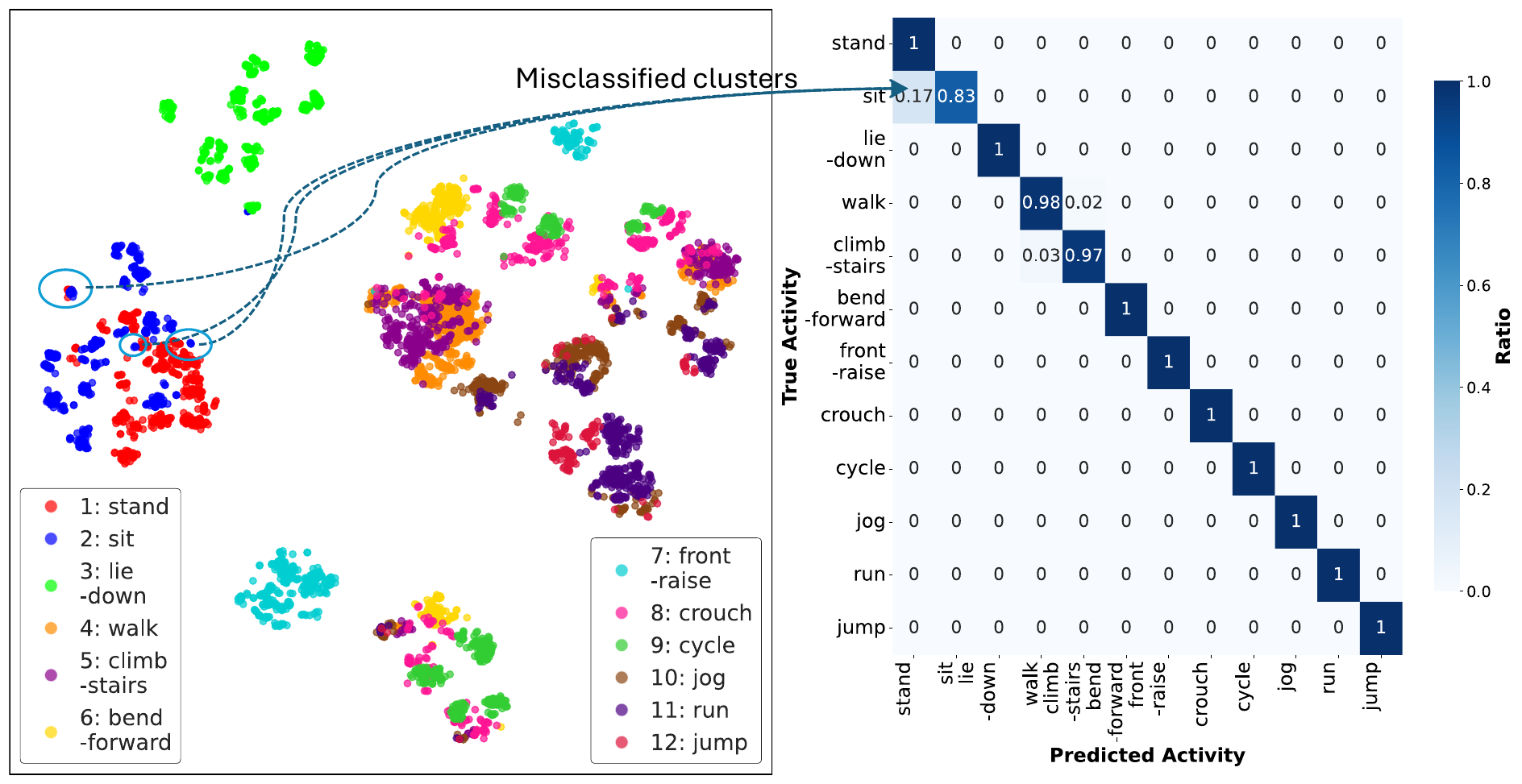} 
    \caption{TSNE analysis of the vector embeddings and confusion matrix of the LLM prediction for MHEALTH dataset.}
    \label{fig:tsne_cm}
\end{figure}

To better understand these retrieval behaviors, we visualize the embeddings using t-SNE in Fig.~\ref{fig:tsne_cm}. The visualization shows that several activity classes (\textit{lie-down}, \textit{stand}, \textit{cycle}, \textit{front-raise}) form compact and well-separated clusters, demonstrating that the embedding stage of our RAG framework encodes rich activity-specific structure. This explains why the retriever succeeds in such cases. At the same time, locomotion activities such as \textit{walk}, \textit{jog}, and \textit{run} display partially overlapping clusters, reflecting the natural difficulty of separating these classes and clarifying why the retriever struggles with them. 

Then, we assess the effect of augmenting retrieval with LLM reasoning. The LLM leverages the retrieved samples as contextual evidence, enabling it to refine predictions, particularly for ambiguous or minority-class instances. 
Overall, Fig.~\ref{fig:ablation-component} shows that incorporating the LLM yields a 14–19\% increase in F1-score across both datasets, highlighting the value of combining retrieval process with language model reasoning. The confusion matrix in Fig.~\ref{fig:tsne_cm} illustrates the resulting \solution{} predictions for MHEALTH dataset. 
We notice that for the majority of the classes LLM predictions are accurate. However, misclassifications appear primarily for the \textit{sitting} activity, consistent with the embedding overlap and isolated points observed in the t-SNE visualization. Although some other clusters exhibit overlap, the absence of isolated points or hard-to-separate samples within these regions enables the LLM to make accurate predictions.

\subsubsection{\textbf{Optimized \solution{}: Prompt optimization}}

As shown in Fig.~\ref{fig:ablation-component}, the integration of the prompt optimizer further enhances accuracy for both MHEALTH and USC-HAD datasets, delivering an additional 2–5\% improvement. The key reason is that systematically refining the \textit{system} prompt allows the optimizer to guide the LLM in leveraging contextual information more effectively, resulting in more accurate classification. For example, Fig.~\ref{fig:mhealth-opt-prompts} shows the top-5 optimized prompts generated for the MHEALTH dataset. We notice that these prompt examples represent some of the best-performing strategies to guide LLM for HAR. 

For instance, they provide explicit guidance for feature extraction, as seen in Prompt 1 and Prompt 2, which detail how to compute mean vectors or energy vectors using statistical measures such as mean\textsuperscript{2}+std\textsuperscript{2}. 
Some prompts, such as Prompt 3 and Prompt 4, guide the LLM on how to process these features and make predictions, for example through conventional ML methods for classification, fusing different measurements through weighted combinations etc.
Overall and most importantly, these prompts break down the task into precise, step-by-step procedures that support the LLM’s reasoning, yielding more consistent and accurate activity classification compared to high-level manual prompts.

We further noticed that the optimized prompts may differ across datasets, as each dataset has unique signal characteristics and activity distributions, requiring the optimizer to adaptively tailor strategies rather than relying on a single prompt solution. In the absence of such an optimizer, discovering high-quality prompts within this vast design space would be ad hoc, highlighting its importance as a critical enabler for reliable RAG-based classification.

\subsubsection{\textbf{Optimized \solution{}: LLM-based Acitivity Descriptor}}

As shown in Fig.~\ref{fig:ablation-component}, integrating this strategy improves \solution{}’s accuracy on the USC-HAD dataset by nearly 6\%. Since text-embedding models are primarily specialized for NLP tasks, providing richer and better-structured descriptions that capture temporal patterns is essential to fully leverage their capabilities, which in turn strengthens the retriever and enhances the overall RAG pipeline.

Fig.~\ref{fig:activity-descriptor} shows example descriptive texts generated for the USC-HAD dataset (sensor at right hip). Unlike basic statistical templates, these descriptions integrate both accelerometer and gyroscope observations and explicitly characterize motion profiles (e.g., periodicity, smoothness, intensity). Beyond raw signal values, the texts embed semantic cues that align with activity type (e.g., rhythmic stepping for walking), thereby enriching the embeddings with motion-specific context that improves retrieval, especially for activities with otherwise similar statistical signatures.

\subsection{Unseen HAR}

To evaluate how LLMs handle unseen activities, we conducted a series of experiments on the PAMAP2 dataset with Baseline \solution{}, focusing on three aspects: (i) the open-set classification capability to detect unseen activities, (ii) the impact of label availability on unseen activity recognition, and (iii) the LLM’s ability to generate meaningful labels for novel activities. We define an \textit{unseen} activity as one that is excluded from the training set (in our case, the vector database).


\subsubsection{\textbf{Open-Set Classification}}

We conducted an open-set classification experiment, following the same train/test splits, segmentation strategy, and unknown proportion definition as in~\cite{lee2022MTMD}. The unknown proportion is defined as the ratio of withheld activity classes to the total number of activity classes, and we evaluated our framework under three openness conditions (30\%, 50\%, and 70\%)  similar to \cite{lee2022MTMD}. Table~\ref{tab:openset-classification} summarizes the results of these experiments, comparing our framework against benchmark baselines under varying openness conditions. Our approach delivers consistently stronger performance, in both lower and higher openness, where \solution{} outperforms MTMD by substantial margins of 3-7\% F1-Score.

\begin{table}[]
\centering
\caption{Open-set classification performance F1-Score (\%).}
\label{tab:openset-classification}
\begin{tabular}{|l|lll|}
\hline
\multicolumn{1}{|c|}{\multirow{2}{*}{\textbf{\textit{Model}}}} & \multicolumn{3}{c|}{\textbf{\textit{Unknown Proportion}}}                                                              \\ \cline{2-4} 
\multicolumn{1}{|c|}{}                                & \multicolumn{1}{c|}{\textbf{{30\%}}} & \multicolumn{1}{c|}{\textbf{{50\%}}} & \multicolumn{1}{c|}{\textbf{{70\%}}} \\ \hline
MTMD~\cite{lee2022MTMD}                                                & \multicolumn{1}{l|}{82.37}         & \multicolumn{1}{l|}{70.33}         & 67.05                              \\ \hline
\textbf{\solution{} (ours)}                                             & \multicolumn{1}{l|}{\textbf{88.75}}              & \multicolumn{1}{l|}{\textbf{77.39}}              & \textbf{69.92}                                    \\ \hline
\end{tabular}
\end{table}

\subsubsection{\textbf{Impact of label availability on unseen activity recognition}}
\label{subsubsec:identifying-unseen}

Having established that \solution{} can effectively detect unseen activities in open-set scenarios,  we next investigate the role of label availability in this process. We adopt a leave-one-class-out protocol: one activity is held out from database indexing and used exclusively as the test set, and this procedure is repeated iteratively for all activities.

We evaluate the impact of label availability by comparing two scenarios:
(i) True label available: The \textit{system} prompt includes the full set of known activity labels along with the true label of the unseen activity (e.g., embeddings from \textit{running} are absent in the database, but the test sample represents \textit{running}, and the prompt still includes \textit{running} as a candidate label).
(ii) True label hidden: The \textit{system} prompt replaces the unseen activity label with the generic placeholder \textit{unseen\_activity} (e.g., in the same example, \textit{running} is replaced by \textit{unseen\_activity}).

These two scenarios affect the label space in two key ways. First, providing the true label introduces a semantic anchor, a meaningful label the LLM has encountered during foundational pretraining (by the vendor), which serves as a natural reference point when the retrieved statistical features align with that activity. 
Second, collapsing all unknown classes into the generic placeholder compresses the label space, increasing the likelihood of borderline cases being pushed toward semantically or statistically similar known classes. 

In our experiment,  when the true label is available, accuracy reaches 78.0 ± 19.8\%, as the model leverages its pretrained knowledge of that activity together with retrieved features. When the true label is replaced with \textit{unseen\_activity}, accuracy drops to 63.2 ± 20.7\%, a decline of nearly 15\% of accuracy. This reduction arises because removing the semantic anchor shifts the model’s predictions toward nearby known classes, producing misclassifications rather than consistently mapping to the generic placeholder. This analysis reveals that LLMs can leverage semantic knowledge embedded in labels to improve unknown activity detection.



\subsubsection{\textbf{Labeling unseen activities}}

Beyond detecting that an activity is unknown, a key advantage of LLM-based approaches is their ability to generate meaningful labels for novel activities.  We formulate unseen-class labelling as a two-stage procedure. The LLM first generates a label for the unseen activity, and second, we compute cosine similarity to map this label to the closest semantically related ground-truth class in the dataset. 
We analyze two cases: holding single and multiple unknown activity classes in the test set.


\begin{figure}
    \centering
    \includegraphics[width=1\linewidth]{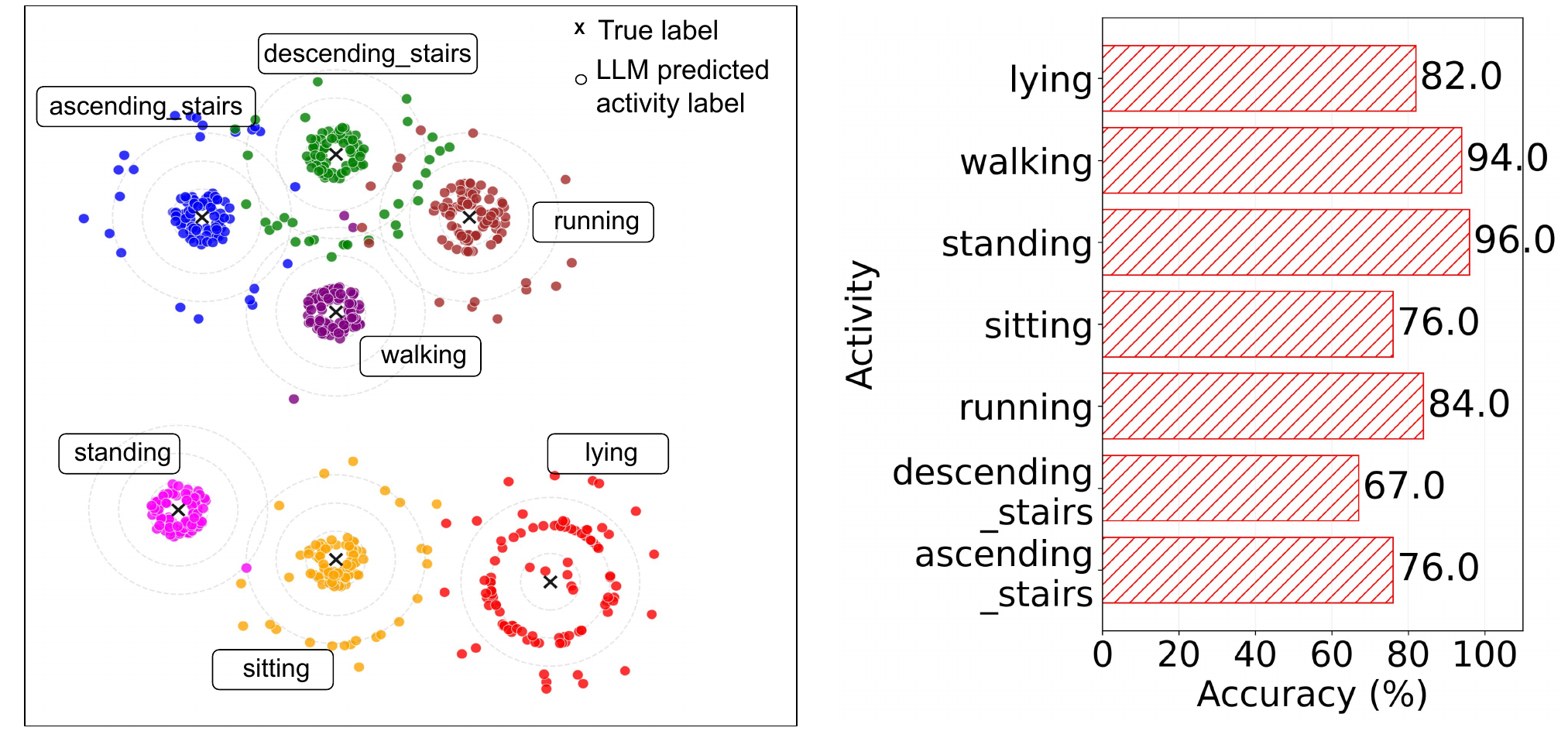}
    \caption{Semantic proximity of LLM-predicted labels to true activities, and class labeling accuracy by unknown activity}
    \label{fig:unseen-activity-analysis}
\end{figure}

\noindent
\textbf{Case 1 - Single unknown activity:}
Fig.~\ref{fig:unseen-activity-analysis} visualizes the semantic proximity of LLM-predicted labels to true activity classes. Each ground-truth activity serves as an anchor (black X), and the predicted labels are plotted as colored points at distances proportional to $(1 - cosine\_similarity)$. Accuracy is then computed by mapping these predicted names to the ground-truth based on semantic similarity.  

Static postures show mixed difficulty: \textit{standing} (96\%) is cleanly separated because of its near-zero motion profile, whereas \textit{sitting} (76\%) is noticeably more error-prone and often drifts toward \textit{lying} (82\%) when trunk orientation is ambiguous. Among locomotion classes, \textit{walking} (94\%) forms a tight, high-confidence cluster, while \textit{running} (84\%) is moderately compact with boundary spillover toward walking at lower speeds/cadences. The stair classes exhibit the most overlap: \textit{ascending\_stairs} (76\%) and \textit{descending\_stairs} (67\%) frequently receive intermediate labels such as “walking upstairs” or generic “stairs,” which are semantically closer to walking and blur the direction cue. This continuous overlap between stair-related activities explains the difficulty of capturing direction-specific cues when both share similar periodic gait dynamics. Overall, the model demonstrates strong reliability in labeling distinct postures and major gait types, while systematic confusion persists among activities that are semantically and physiologically related.



\noindent
\textbf{Case 2 - Multiple unknown activities:}
Conventional DL models trained on fixed taxonomies have limited ability to handle novelty. They typically extend the label space in a binary way, sending all out-of-distribution inputs to a single \textit{unknown} class. While this provides a basic rejection mechanism, it collapses diverse forms of novelty into one undifferentiated bucket, masking important differences between unseen behaviors. In contrast, LLMs leverage pretrained semantic knowledge to assign more fine-grained and human-interpretable labels, effectively partitioning the unknown space into meaningful subcategories. This preserves semantic richness that fixed-taxonomy systems usually discard, enabling a more informative treatment of unseen activities.

To probe this capability, we extend our experiments by withholding multiple novel classes simultaneously. This setting requires the model to differentiate between distinct unknowns. We further vary the number of withheld classes to evaluate scalability in the multi-unknown regime.  \solution{} achieves a best accuracy of 96.47\% when a single class is withheld, and maintains strong performance with 92.73\% for two unknown classes and 88.90\% for three, demonstrating its robustness in generating consistent and separable labels even under increasing ambiguity.
These results demonstrate that \solution{} is not only competitive in moderate open-set settings but also robust when the unknown space becomes larger and more challenging, underscoring \solution{}'s effectiveness for real-world open-world recognition scenarios.

\section{Discussion and Limitations}
\label{sec:discussion}

\subsection{Limitations}

The \solution{} framework demonstrates that RAG with LLMs can be effective for HAR, but some limitations remain. First, sensor windows (e.g., 60 channels in the Skoda) are represented with statistical descriptors to fit within LLM context limits, which sacrifices fine-grained temporal patterns and inter-channel dependencies. Future work may incorporate raw or downsampled sequences using techniques like PCA or adaptive sampling. Second, we relied on a general-purpose text embedding model, requiring statistical-to-text conversion. While workable, this limits the expressiveness of sensor readings; future advances in general-purpose time-series embedding models—analogous to CLIP for images—could allow direct indexing of raw windows without preprocessing.

\subsection{Future work}

Beyond these limitations, several directions highlight the broader potential of our approach. One avenue is extending \solution{} to a multi-agent framework, where specialized agents handle different sensors or body parts, each with its own RAG module. By collaborating, these agents could resolve ambiguities and scale to complex multi-sensor settings. Another consideration is model choice: we selected gpt-5-mini as our base model to balance cost and reasoning capability. While different models may require distinct prompts for optimal performance—a direction we leave for future work—our prompt optimization methodology (\S~\ref{subsec:promptopt}) can automatically identify effective prompts across LLMs. Finally, while LLMs exhibit some non-determinism, our retrieval-augmented design mitigates this by grounding predictions in activity examples, yielding stable classifications across multiple runs.

\subsection{Cost, Time, and Deployment Complexity}
A core strength of \solution{} is that it delivers SOTA accuracy while remaining computationally lightweight, training-free, and highly cost-effective. In contrast to deep learning based HAR systems, which require lengthy training cycles and repeated fine-tuning for each dataset, \solution{} performs inference using only basic statistical descriptors, a vector database retrieval, and a single LLM call per window. This shift from model-centric training to retrieval-augmented reasoning significantly reduces both engineering overhead and operational cost.

\subsubsection{Cost Efficiency}
\solution{} incurs only minimal cost across all stages of the pipeline. On the MHEALTH dataset, generating embeddings for the 4,000-sample indexing set cost \$0.0176, and the optional prompt optimization procedure, which is run once offline during development, cost \$12.56 and does not scale with dataset size. During deployment, the only recurring expense is the LLM inference step. Running 600 \solution{} predictions costs \$0.3739, which gives an amortized per-sample inference cost of \$0.000623. These values show that \solution{} operates at a very small fraction of the cost associated with deep learning pipelines that typically require multi-epoch GPU training and periodic fine-tuning.

\subsubsection{Time Efficiency}
\solution{} also offers a lightweight inference-time profile since each prediction requires only a single embedding lookup and a single LLM forward pass. Unlike deep learning pipelines, which require multi-hour training, repeated fine-tuning, and multi-seed runs to achieve stable performance, \solution{} avoids all training time entirely. At inference time, prediction latency is dominated by a single LLM call, making the system comparable to the time of one forward pass in a conventional neural classifier without the cost of training that classifier in the first place.

\subsubsection{Deployment and Adaptation Simplicity}
Although \solution{} contains several components, the overall system is significantly simpler to deploy and maintain than deep learning based HAR approaches. All activity knowledge is stored in the vector database instead of being encoded in model parameters. Adapting the system to new users or adding unseen activities requires only computing statistical descriptors and inserting the corresponding embeddings into the database. This avoids the need for gradient updates or retraining and shifts the computational burden from heavy iterative optimization to lightweight descriptor extraction, fast retrieval, and a single LLM call. As a result, \solution{} provides a practical and scalable solution for real-world settings where users, sensors, and activities evolve.
\section{Conclusion}
\label{sec:conclusion}

This paper introduced \solution{}, a training-free HAR framework that combines RAG with LLM reasoning. Activity windows are represented by statistical descriptors, indexed in a vector database, and classified through similarity search and LLM reasoning, achieving SOTA results without any training or fine-tuning. Beyond closed-set recognition, \solution{} demonstrated open-set reasoning, where the LLM inferred unseen activities using its prior knowledge—an ability traditional DL models lack. Overall, by uniting retrieval, LLM reasoning and other optimization strategies, \solution{} provides a scalable, training-free alternative to conventional HAR pipelines, with strong potential for real-world deployment and future research in pervasive computing and retrieval-augmented inference.

\bibliographystyle{IEEEtranS}
\balance
\bibliography{99_bib}

\end{document}